\documentclass[journal]{IEEEtran}
\usepackage[colorlinks,linkcolor=red]{hyperref}

\usepackage{amsmath}
\usepackage{epsfig}
\usepackage{graphicx}
\usepackage{subfigure}
\usepackage{multirow}
\usepackage{color}
\usepackage{changepage}
\usepackage{times}
\usepackage{float}
\usepackage{amssymb}
\usepackage{multirow}
\usepackage{bbding}
\usepackage{booktabs}
\usepackage{algorithm}
\usepackage{algpseudocode}
\usepackage{caption2}
\usepackage{verbatim}
\usepackage{dsfont}
\usepackage{amssymb}
\usepackage{stfloats}
\usepackage{tikz}
\usetikzlibrary{arrows}

\usepackage{url}
\ifCLASSINFOpdf
\else
\fi

\begin{document}
%
\title{Masked Face Recognition Dataset and Application}
%
%
%

\author{Zhongyuan~Wang, Guangcheng~Wang, Baojin~Huang, Zhangyang~Xiong, Qi~Hong, Hao~Wu, Peng~Yi, Kui~Jiang, Nanxi~Wang, Yingjiao~Pei, Heling~Chen, Yu~Miao, Zhibing~Huang, and~Jinbi~Liang
\thanks{Z. Wang, G. Wang, B. Huang, Z. Xiong, Q. Hong, H. Wu, P. Yi, K. Jiang, N. Wang, Y. Pei, H. Chen, Y. Miao, Z. Huang, and J. Liang are with the National Engineering Research Center for Multimedia Software, School of Computer Science, Wuhan University, Wuhan 430072, China. \emph{(Corresponding author: Zhongyuan Wang, wzy\_hope@163.com).}}
}

%
%

\maketitle

\begin{abstract}
In order to effectively prevent the spread of COVID-19 virus, almost everyone wears a mask during coronavirus epidemic. This almost makes conventional facial recognition technology ineffective in many cases, such as community access control, face access control, facial attendance, facial security checks at train stations, etc. Therefore, it is very urgent to improve the recognition performance of the existing face recognition technology on the masked faces. Most current advanced face recognition approaches are designed based on deep learning, which depend on a large number of face samples. However, at present, there are no publicly available masked face recognition datasets. To this end, this work proposes three types of masked face datasets, including Masked Face Detection Dataset (MFDD), Real-world Masked Face Recognition Dataset (RMFRD) and Simulated Masked Face Recognition Dataset (SMFRD). Among them, to the best of our knowledge, RMFRD is currently the world's largest real-world masked face dataset. These datasets are freely available to industry and academia, based on which various applications on masked faces can be developed. The multi-granularity masked face recognition model we developed achieves 95\% accuracy, exceeding the results reported by the industry. Our datasets are available at: \url{https://github.com/X-zhangyang/Real-World-Masked-Face-Dataset}.
\end{abstract}


\IEEEpeerreviewmaketitle

\section{Background}
\IEEEPARstart{A}{lmost} everyone wears a mask during the COVID-19 coronavirus epidemic. Face recognition techniques, the most important means of identification, have nearly failed, which has brought huge dilemmas to authentication applications that rely on face recognition, such as community entry and exit, face access control, face attendance, face gates at train stations, face authentication based mobile payment, face recognition based social security investigation, etc. In particular, in the public security check like railway stations, the gates based on traditional face recognition systems can not effectively recognize the masked faces, but removing masks for  passing authentication will increase the risk of virus infection. Because the COVID-19 virus can be spread through contact, the unlocking systems based on passwords or fingerprints are unsafe. It is much safer through face recognition without touching, but the existing face recognition solutions are no longer reliable when wearing a mask. To solve above mentioned difficulties, it is necessary to improve the existing face recognition approaches that heavily rely on all facial feature points, so that identity verification can still be performed reliably in the case of incompletely exposed faces.

The state-of-the-art face recognizers are all designed based on deep learning, which depend on massive training dataset~[\ref{ref:arcface}]-[\ref{ref:8099646}]. Thus, developing face recognition algorithms for masked faces requires a large number of masked face samples. At present, there is no publicly available masked face dataset, and so this work proposes to construct masked face datasets by different means.

\section{Proposed Datasets}
Regarding the current popular face masks, there are two closely related and different applications, namely,facial mask detection task and masked face recognition task. Face mask detection task needs to identify whether a person wear a mask as required. Masked face recognition task needs to identify the specific identity of a person with a mask. Each task has different requirements for the dataset. The former only needs masked face image samples, but the latter requires a dataset which contains multiple face images of the same subject with and without a mask. Relatively, datasets used for the face recognition task are more difficult to construct.

\begin{figure}[!t]
\vspace{0.1cm}
\center
\small
\includegraphics[width=0.985\linewidth]{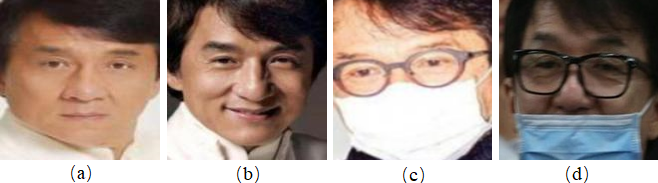}\\
\vspace{-0.1cm}
\caption{Examples of a pair of face images. (a) and (b) are normal face images. (c) and (d) are masked face images.}
\label{fig:01}
\vspace{0.1cm}
\end{figure}

In order to handle masked face recognition task, this paper proposes three types of masked face datasets, including Masked Face Detection Dataset (MFDD), Real-world Masked Face Recognition Dataset (RMFRD) and Simulated Masked Face Recognition Dataset (SMFRD). The introduction of MFDD, RMFRD and SMFRD is shown below.

\begin{itemize}[\IEEEsetlabelwidth{Z}]
\vspace{0.1cm}
\item[$\bullet$] MFDD: The source of MFDD mainly includes two parts: (a) Some of samples are from related researches~[\ref{ref:AIZOOM}]; (b) The other part of MFDD is crawled from the Internet. We further label the crawled face images, performing annotations such as whether the face wears a mask and the position coordinates of the masked faces. This built dataset contains 24,771 masked face images. MFDD dataset can be used to train an accurate masked face detection model, which serves for the subsequent masked face recognition task.
Additionally, it can also be used to determine whether a person is wearing a mask, as it is illegal without wearing a mask during coronavirus epidemic.

\vspace{0.1cm}
\item[$\bullet$] RMFRD: A python crawler tool is used to crawl the front-face images of public figures and their corresponding masked face images from massive Internet resources. Then, we manually remove the unreasonable face images resulting from wrong correspondence. The process of filtering images takes a lot of manpower. Similarly, we crop the accurate face areas with the help of semi-automatic annotation tools, like LabelImg and LabelMe~[\ref{ref:Label-tool}]. The dataset includes 5,000 pictures of 525 people wearing masks, and 90,000 images of the same 525 subjects without masks. To the best of our knowledge, this is currently the world's largest real-world masked face dataset. Fig. 1 shows pairs of facial image samples.

\vspace{0.1cm}
\item[$\bullet$] SMFRD: In order to expand the volume and diversity of the masked face recognition dataset, we meanwhile have taken alternative means, which is to put on masks on the existing public large-scale face datasets.
To improve data manipulation efficiency, we have developed a mask wearing software based on Dlib library~[\ref{ref:Dlib-library}] to perform mask wearing automatically. This software is then used to wear masks on face images in the popular face recognition datasets, presently including LFW~[\ref{ref:lfw}] and Webface~[\ref{ref:webface}] datasets. This way, we additionally constructed a simulated masked face dataset covering 500,000 face images of 10,000 subjects.
In practice, the simulated masked face datasets can be used along with their original unmasked counterparts. Fig. 2 shows a set of simulated masked face images.

\vspace{0.1cm}
\end{itemize}

\begin{figure}[!t]
\vspace{0.1cm}
\center
\small
\includegraphics[width=0.985\linewidth]{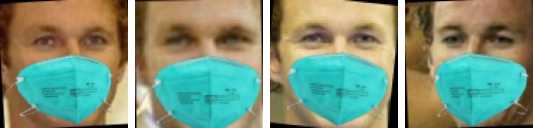}\\
\vspace{-0.1cm}
\caption{Samples of a set of simulated masked face images.}
\label{fig:01}
\vspace{0.1cm}
\end{figure}

\section{Masked Face Recognition}
Face-based identification can be roughly divided into two application scenarios: uncontrolled and controlled application environments. The former mainly refers to public video surveillance situations, where face shooting distance, view of sight, pose, occlusion and lighting are all uncertain. In these cases, the accuracy of face recognition is relatively low. Moreover, the accuracy will be further reduced when wearing a face mask. However, there are also a large number of controlled application scenarios, such as attendance checks in work places, security checks at train stations and facial scan payments, etc. In these situations, subjects are usually in a cooperative manner, typically, approaching and facing up the camera. Thus high-quality frontal face images are readily acquired, so that the masked face recognition task is no longer so difficult. Even if the mask covers part of the face, the features of upper half of the face, such as eye and eyebrow, can still be used to improve the availability of the face recognition system. Of course, the premise is to exclude mask interference and give higher priority to useful exposed face features.

Our proposed masked face recognition technique has been blessed with two aspects. One is the built dataset, and the other is the full use of uncovered useful face features. We took advantages of the existing public face recognition datasets, and combined them with the self-built simulated masked faces as well as the masked faces from actual scenes as the final dataset to train a face-eye-based multi-granularity recognition model. In particular, we applied different attention weights to the key features in visible parts of the masked face, such as face contour, ocular and periocular details, forehead, and so on, which effectively addresses the problem of uneven distribution of facial discriminative information. As result, we promote the recognition accuracy of masked faces from the initial 50\% to 95\%.

\section{Application Status and Prospect}
Probably because of the sudden emergence of the COVID-19 epidemic, at present,
there are few institutions that apply facial recognition technology to people wearing masks. Based on our survey, Sense Time Technology reported a pass rate of 85\% when the person exposes 50\% of the nose~[\ref{ref:report}]. Hanvon Technology also reported that the accuracy of masked face recognition is about 85\%~[\ref{ref:Hanvon}]. The best result reported so far is from MINIVISION Technology, with an accuracy of over 90\%~[\ref{ref:MINIVISION}]. Our face-eye-based multi-granularity model achieves 95\% recognition accuracy. Generally, masked face recognition technology can be used to identify people wearing masks but it is still not very reliable compared to the regular facial recognition technology which already witnessed an accuracy of over 99\%. Another related task is face mask recognition, that is, identifying whether a person is wearing a mask as required or not. Because the task is relatively simple, the recognition accuracy is much higher. Tencent, Baidu, and Jingdong all reach a recognition accuracy of more than 99\%.

We built the MFDD, RMFRD and SMFRD datasets, and developed a state-of-the-art algorithm based on these datasets. The algorithm will serve the applications of contactless face authentication in community access, campus management, and enterprise resumption scenarios. Our research has contributed scientific and technological power to the prevention and control of coronavirus epidemics and the resumption of production in industry. Furthermore, due to the frequent occurrence of haze weather, people will often wear masks, and the need for face recognition with masks will persist for a long time.

%


\appendices
%


\section*{Acknowledgment}

This work was supported by National Natural Science Foundation of China (U1903214, U1736206, 61971165, 61502354), Hubei Province Technological Innovation Major Project (2019AAA049,2019AAA045), and The Outstanding Youth Science and Technology Innovation Team Project of Colleges and Universities in Hubei Province (T201923).

Because the university was closed during the coronavirus epidemic, we thank the leaders and colleagues for coordinating research resources. Part of the computing has been done on the supercomputing system in the Supercomputing Center of Wuhan University.

\ifCLASSOPTIONcaptionsoff
  \newpage
\fi


\begin{thebibliography}{1}

\bibitem{33}\label{ref:arcface}
J. Deng, J. Guo, N. Xue, S. Zafeiriou, ``ArcFace: Additive Angular Margin Loss for Deep Face Recognition,'' in \emph{CVPR}, Jun. 2019, pp. 4685-4694.

\bibitem{33}\label{ref:Fair-LOSS}
B. Liu, W. Deng, Y. Zhong, M. Wang, J. Hu, X. Tao, and Y. Huang, ``Fair Loss: Margin-Aware Reinforcement Learning for Deep Face Recognition'', in \emph{ICCV}, Oct. 2019, pp. 10051-10060.

\bibitem{33}\label{ref:Sphereface}
W. Liu, Y. Wen, Z. Yu, M. Li, B. Raj, and L. Song, ``Sphereface: Deep hypersphere embedding for face recognition'',in \emph{CVPR}, Jul. 2017, pp. 6738-6746.

\bibitem{33}\label{ref:liu2016large}
W. Liu, Y. Wen, Z. Yu, and M. Yang, ``Large-margin softmax loss for convolutional neural networks'', in \emph{ICML}, 2016, pp. 507-516.

\bibitem{33}\label{ref:8099646}
A. T. Tran, T. Hassner, I. Masi, and G. Medioni, ``Regressing Robust and Discriminative 3D Morphable Models with a Very Deep Neural Network'', in \emph{CVPR}, Jul. 2017, pp. 1493-1052.

\bibitem{33}\label{ref:AIZOOM}
\url{https://zhuanlan.zhihu.com/p/107719641?utm_source=com.yinxiang}

\bibitem{33}\label{ref:Label-tool}
\url{https://tzutalin.github.io/labelImg/}

\bibitem{33}\label{ref:Dlib-library}
\url{http://dlib.net/}

\bibitem{33}\label{ref:lfw}
G. B. Huang, M. Mattar, T. Berg, and E. Learned-Miller, ``Labeled faces in the wild: A database for studying face recognition in unconstrained environments'', Technical report, 2007.

\bibitem{33}\label{ref:webface}
D. Yi, Z. Lei, S. Liao, and S. Z. Li, ``Learning face representation from scratch'', \emph{arXiv:1411.7923}, 2014.

\bibitem{33}\label{ref:report}
\url{http://ai.cps.com.cn/article/202002/937650.html}

\bibitem{33}\label{ref:Hanvon}
\url{https://baijiahao.baidu.com/s?id=1658872342983093939&wfr=spider&for=pc}

\bibitem{33}\label{ref:MINIVISION}
\url{https://blog.csdn.net/aizhushou/article/details/104393844}

\end{thebibliography}
\end{document}